%% file: main.tex
\documentclass[runningheads]{llncs}
\usepackage{graphicx}
\usepackage{booktabs} 
\usepackage{verbatim}
\usepackage{multirow}
\usepackage{longtable}
\usepackage{array}
\usepackage{subfigure}
\usepackage{stfloats}
\usepackage{balance}
\usepackage{multicol}
\usepackage{color}
\usepackage{bm}
\usepackage{amsmath}
\usepackage{amssymb}
\usepackage{booktabs} 
\usepackage{epsfig}
\usepackage{enumitem}
\usepackage{cleveref}
\usepackage{arydshln}
\usepackage{color}
\usepackage[linesnumbered, ruled, vlined]{algorithm2e}

\newcommand{\hide}[1]{} 
\newcommand{\vpara}[1]{\vspace{0.1in}\noindent\textbf{#1 }}
\newcommand{\ipara}[1]{\vspace{0.1in}\noindent\textit{#1 }}

\newcommand{\beq}[1]{\vspace{-0.02in}\begin{equation}#1\end{equation}\vspace{-0.02in}}
\newcommand{\beqn}[1]{\vspace{-0.03in}\begin{eqnarray}#1\end{eqnarray}\vspace{-0.03in}}

\newcommand{\model}{JarKA}
\newcommand{\smodel}{JarKA\space}

\newcommand{\samodel}{JarKA-r\space}

\newcommand{\srmodel}{JarKA-a\space}

\begin{document}
%
\title{\model: Modeling Attribute Interactions for Cross-lingual Knowledge Alignment}
\titlerunning{JarKA}
%

\author{Bo Chen\inst{1,2} \and
Jing Zhang$^*$\inst{1,2} \and
Xiaobin Tang\inst{1,2} \and 
Hong Chen\inst{1,2} \and
Cuiping Li\inst{1,2}}
\authorrunning{Chen et al.}
%
\institute{Key Laboratory of Data Engineering and Knowledge Engineering of Ministry of Education, Renmin University of China \and Information School, Renmin University of China\\
\email{\{bochen, zhang-jing, txb, chong, licuiping\}@ruc.edu.cn}}

\maketitle              

\input{abstract.tex}


\input{intro.tex}

\input{problem.tex}

\input{approach}

\input{exp}

\input{conclusion}

\balance

\bibliographystyle{abbrv}
\bibliography{references}

\end{document}

%% file: abstract.tex
\begin{abstract}


Cross-lingual knowledge alignment is the cornerstone in building a comprehensive knowledge graph (KG), which can benefit various knowledge-driven applications. As the structures of  KGs are usually sparse, attributes of entities may play an important role in aligning the entities.  However, the heterogeneity of the attributes across KGs prevents from accurately embedding and comparing entities. To deal with the issue, we propose to model the interactions between attributes, instead of globally embedding an entity with all the attributes. We further propose a joint framework to merge the alignments inferred from the attributes and the structures. Experimental results show that the proposed model outperforms the state-of-art baselines by up to 38.48\% HitRatio@1. The results also demonstrate that our model can infer the alignments between attributes, relationships and values, in addition to entities.

\end{abstract}

%% file: intro.tex
\section{Introduction}
\label{sec:intro}

DBpedia, Freebase, YAGO and so on have been published as noteworthy large and freely available knowledge graphs (KGs), which can benefit many knowledge-driven applications.
However, the knowledge embedded in different languages is extremely unbalanced. For example, DBpedia contains about 2.6 billion triplets in English, but only 889 million and 278 million triplets in French and Chinese respectively. 
Creating the linkages between cross-lingual KGs can reduce the gap of acquiring knowledge across multiple languages and benefit many applications such as machine translation, cross-lingual QA and cross-lingual IR.

\begin{figure}[t]
	\centering
	\includegraphics[width=\textwidth]{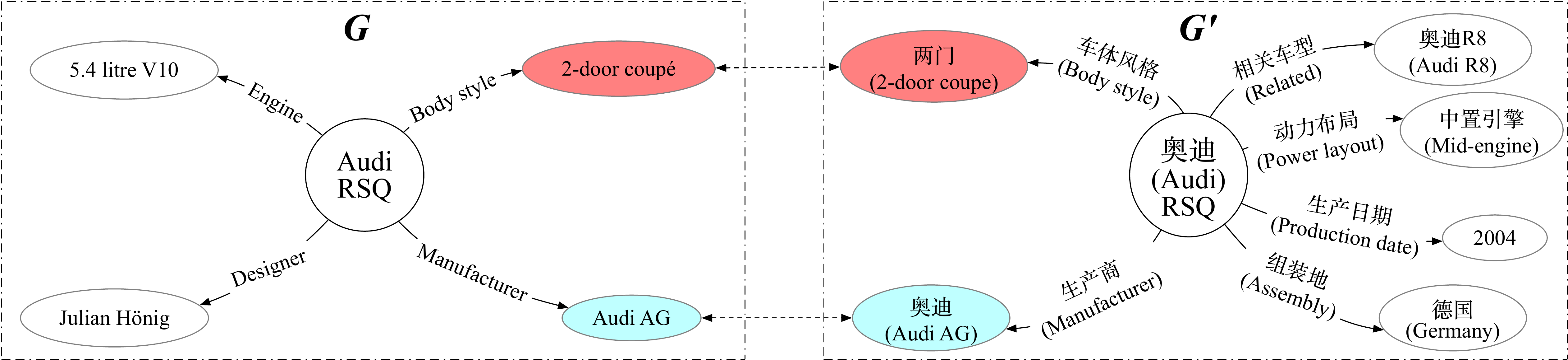}
	\caption{\label{fig:motivation} Illustration of different attributes of same entities in two cross-lingual knowledge graphs from wikipedia.}
\end{figure}

Recently, much attention has been paid to leveraging the embedding techniques to align entities between two KGs. Some of them only leverage the structures of the KGs, i.e., the relationship triplets in the form of $\langle$entity, relationship, entity$\rangle$ to learn the structure embeddings of entities ~\cite{cao2019multi,hao2016joint,sun2018bootstrapping}. 
However, the structures of some KGs are sparse, making it difficult to learn the structure embeddings accurately.  
Other efforts are made to incorporate the attribute triplets in the form of  $\langle$entity, attribute, value$\rangle$ to learn the attribute embeddings of entities~\cite{sun2017crosslingual,Trsedya2019,wang2018cross,zhang2019multi}. For example,
JAPE~\cite{sun2017crosslingual} embeds attributes via attributes' concurrence. Wang et al.~\cite{wang2018cross} adopt GCNs to embed entities with the one-hot representations of the attributes. Trsedya et al.~\cite{Trsedya2019} and MultiKE~\cite{zhang2019multi} embed the literal values of the attributes.
Despite the existing studies on incorporating the attribute triplets to align entities, there are still unsolved challenges. 

\vpara{Challenge 1: Heterogeneity of Attributes.} Different KGs may hold heterogeneous attributes, resulting in the difficulty of aligning entities.  For example, in Fig. \ref{fig:motivation}, two entities from cross-lingual KGs named ``Audi RSQ" are the same entity. Although the attributes ``Manufacturer" and ``Body style" and their values in English correspond to certain attribute triplets in Chinese, there are still many attribute  such as ``Designer" and ``Engine" in English that cannot find any counterpart in Chinese. However, if we embed an entity by all its attribute triplets and then compare two entities by their attribute embeddings \cite{Trsedya2019,zhang2019multi}, the effects of the same attribute triplets will be diluted by other different ones. 

\vpara{Challenge 2: Multi-view Combination.} To combine the effects from attributes and structures, existing works usually learn a combined embedding for each entity, based on which they infer the alignments. For example,  JAPE~\cite{sun2017crosslingual} and AttrE~\cite{Trsedya2019} refine the structure embeddings by the closeness of the corresponding attribute embeddings. MultiKE~\cite{zhang2019multi} map the attribute and structure embeddings into a unified space. However, the issue of the missing attributes or relationships triplets may result in the inaccurate attribute or structure embeddings, which will propagate the errors to the combined embeddings. 

Besides the above two challenges, most of the existing works~\cite{cao2019multi,sun2017crosslingual,zhang2019multi} only focus on aligning entities, or at most  relationships, but ignore  attributes and values.  However, the alignment of different objects influence each other. A unified way to align all of these objects simultaneously is worth studying.

\vpara{Solution.}
To deal with the above challenges, we propose a joint model --- \smodel to \underline{J}ointly model the \underline{a}ttributes interactions and \underline{r}elationships for cross-lingual \underline{K}nowledge  \underline{A}lignment. The two views are carefully merged to reinforce the training performance iteratively.
The contributions can be summarized as:

\begin{itemize}
	\item We comprehensively formalize cross-lingual knowledge alignment as  linking  entities, relationships, attributes and values across cross-lingual KGs.
	
	\item To tackle the first challenge, we propose an interaction-based attribute model to capture the attribute-level interactions  between two entities instead of globally representing the two entities. A matrix-based strategy is further proposed to accelerate the similarity estimation. 

	\item To deal with the second challenge, we propose a joint framework to combine the alignments inferred by the attribute model and relationship model respectively instead of learning a combined embedding. Three different merge strategies are proposed to solve the conflicting alignments.

\item 
Experimental results on several datasets of cross-lingual KGs demonstrate that  \smodel   significantly outperforms  state-of-the-art comparison methods (improving 2.35-38.48\% in terms of Hit Ratio@1). 
	
\end{itemize}

%% file: problem.tex
\section{Problem Definition}
\label{sec:problem}

\begin{definition}
\textbf{Knowledge Graph}:	We denote the KG as union of the \textbf{relationship triplets} and the \textbf{attribute triplets}, i.e., $G=\{ ( h,r,t )\} \cup \{( h,a, v)\}$, where $(h,r,t)$ is a relationship triplet consisting of a head entity $h$, a relationship $r$, and a tail entity $t$, and $(h, a, v)$ is an attribute triplet consisting of a head entity $h$, an attribute $a$ and its value $v$. We also use $e$ to denote entity.
\end{definition}


We distinguish the two kinds of triplets as they are independent views that can take different effects on alignment.

\begin{problem}
\textbf{Cross-lingual Knowledge Alignment}: Given two cross-lingual KGs $G$ and $G'$, and the seed set $I$ of the aligned entities, relationships, attributes, and values, i.e., $I = \{(e\sim e')\} \cup \{(r \sim r')\} \cup \{(a \sim a')\} \cup \{(v \sim v')\}$ \footnote{Please refer to section 4 for how to obtain $I$.},the goal is to augment $I$ by the inferred new alignments between $G$ and $G'$.
\end{problem}


%% file: approach.tex
\section{\smodel Model}
\label{sec:rl}

We propose an interaction-based attribute model to leverage the $(h, a, v)$ triplets, an embedding-based relationship model to leverage the $\{(h,r,t)\}$ triplets, and then incorporate the two models by a carefully designed joint framework.

\subsection{Interaction-based Attribute Model}
Existing methods represent an entity globally by all its associated $(h, a, v)$ and then compare the entity embedding between entities~\cite{Trsedya2019,zhang2019multi}. However, as shown in Fig.~\ref{fig:motivation}, two entities from cross-lingual KGs may have heterogeneous attribute. The irrelevant attribute triplets between two entities may dilute the effects of their similar attribute triplets if globally embedding the entities. 

\begin{figure}[t]
 	\centering
 	\includegraphics[width=\textwidth]{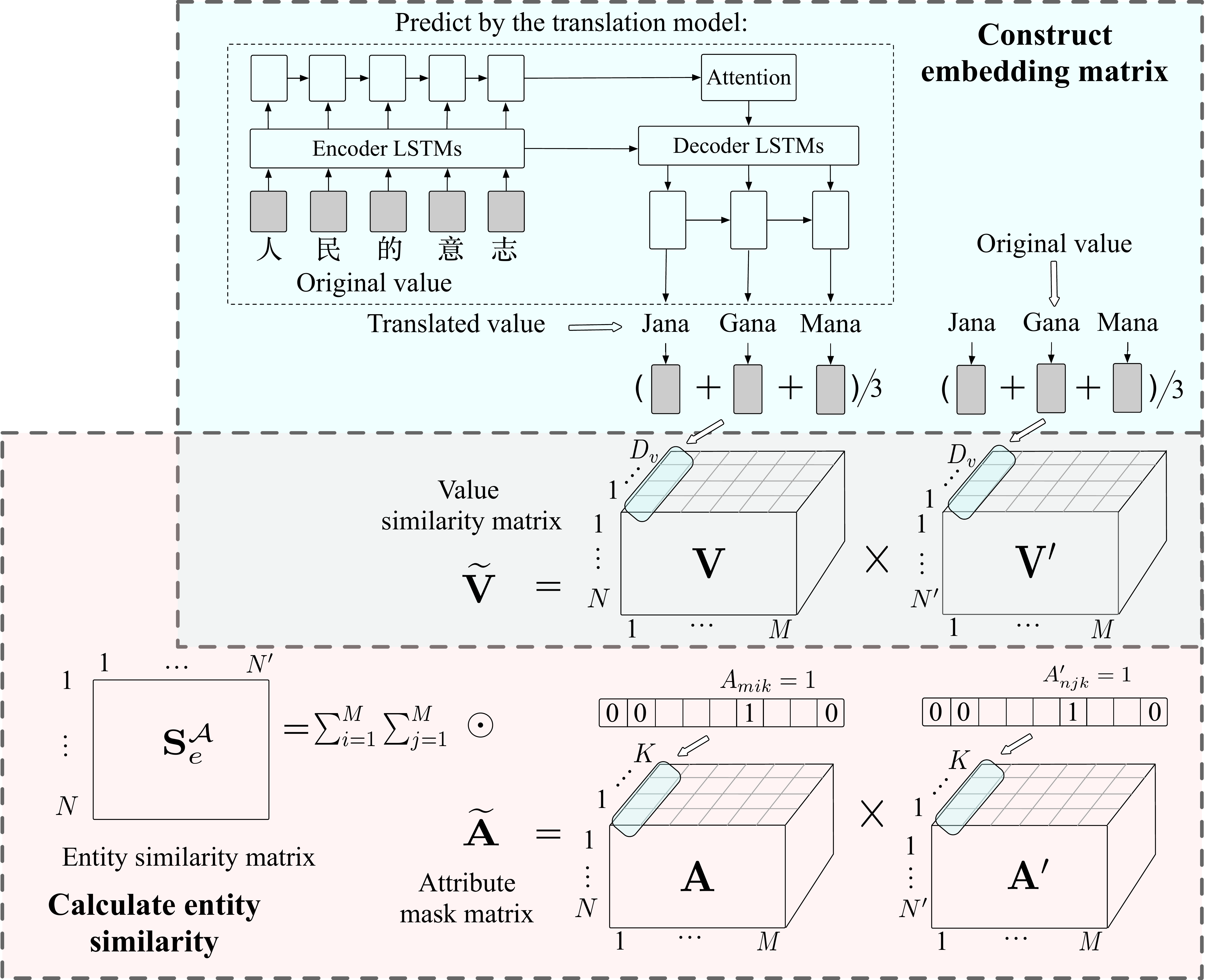}
 	\caption{\label{fig:EALmodel} Illustration of the proposed attribute model. $V$ and $V'$ are the value embedding matrices, and $A$ and $A'$ are the attribute identification matrices for G and G' respectively. The figure can be read from left to right and top to bottom.}
 \end{figure}

To deal with the above issue, we propose an interaction-based attribute model to directly estimate the similarity of two entities by capturing the interactions between their attributes and values. The model mimics the process that humans solve the problem. The humans usually align two entities if they have many same attributes with same values. Following this, we firstly find all the aligned attribute pairs of two entities, and then compare its values.
Since the number of  the attributes are far smaller than that of the values in KGs, we initialize the aligned attributes by the attribute seed pairs and gradually extend them by our joint framework, which will be introduced in the following section. To compare the large number of cross-lingual values, we train a machine translation model and use it to estimate the BLEU score~\cite{papineni2002bleu} of two cross-lingual values as their similarity. 
Unfortunately, following the idea, we need to enumerate and invoke the translation model for maximal $M$ attribute pairs for each entity pair, resulting in $O(N \times N' \times M)$ time complexity when there are $N$ and $N'$ entities in $G$ and $G'$ respectively, which is too inefficient to finish within available time.
To accelerate the similarity estimation, we represent each knowledge graph as a 3-dimension value embedding matrix and then perform an efficient matrix-based strategy to calculate entity similarities. Figure~\ref{fig:EALmodel} illustrates the whole process of the proposed attribute model. In the following part, we will explain the details.

\vpara{Embed Cross-lingual Attribute Values.}
We build an neural machine translation model (NMT)~\cite{bahdanau2015neural}  to capture semantic similarities between cross-lingual values. 
We pretrain NMT based on the value seeds\footnote{In the future, external cross-lingual corpus can be easily used to pre-train the model.}. Since the seeds are limited, we will update NMT by the newly discovered value seeds iteratively. 

Then we use NMT to project cross-lingual values into the same vector space.  Specifically, for each attribute of each entity in $G$, we first invoke NMT to predict the translated value given its original value, and then look up the word embedding for each word in the translated value. While for each attribute of each entity in $G'$, we directly look up the word embedding for each word in its original  value. With the help of NMT, the embeddings of the cross-lingual values can be unified in the same space. Then we average all the word embeddings in the value as its value embedding. The dimension is denoted as $D_v$.
 
\vpara{Estimate Entity Similarities by Matrix-based Strategy.}
We construct a 3-dimension value embedding matrix $\mathbf{V} \in  \mathbb{R}^{N \times M \times D_v}$ for $G$ and a similar matrix $\mathbf{V}' \in  \mathbb{R}^{N' \times M \times D_v}$ for $G'$, where each element $\textbf{V}_{mi}$ indicates the $i$-th value embedding of the $m$-th entity.
Then we use the einsum operation
\beq{
	\label{eq:einsum}
	\text{einsum}(NMD_v, N'MD_v \rightarrow NN'MM),
}
\noindent  i.e., Einstein summation convention~\cite{ahlander2002einstein}, to make a multi-dimensional matrix product of $\mathbf{V}$ and $\mathbf{V}'$ to obtain the value similarity matrix $\widetilde{\mathbf{V}} \in \mathbb{R}^{N \times N' \times M \times M}$. 

What's more, it is unnecessary to compare the values of different attributes. For example, although the attributes ``birthplace" and ``deathplace" have the same value ``New York", they cannot reflect the similarity of two entities. So, we build an attribute mask matrix $\widetilde{\mathbf{A}}$ to limit the computation within the values of the aligned attributes. Specifically, we  prepare a 3-dimension attribute identification matrix $\mathbf{A} \in  \mathbb{R}^{N \times M \times K}$ for $G$ and $\mathbf{A}' \in  \mathbb{R}^{N' \times M \times K}$ for $G'$, where $K$ denotes the number of  the united frequent attributes in $G$ and $G'$. Each row in $\mathbf{A}$ or $\mathbf{A}'$ is an one-hot vector, with an element $A_{mik}\!=\!1$ if the $i$-th value of the $m$-th entity belongs to the $k$-th attribute, and $A_{mik}\!=\!0$ otherwise. Note the one-hot identification vector depends on the existing aligned attributes, which will be gradually extended with the joint model iteratively. 
Whenever two attributes are discovered to be aligned, we will unify their identification. For example, when the $k$-th attribute in $G$ and the $t$-th attribute in $G'$ are aligned, we replace the identification $k$ with $t$, i.e., any row with $A_{mik}=1$ will be changed to $A_{mit}=1$. 
Then we multiply $\mathbf{A}$ and $\mathbf{A}'$ in the same way as Eq.\eqref{eq:einsum} to obtain an attribute mask matrix $\widetilde{\mathbf{A}} \in \mathbb{R}^{N \times N' \times M \times M}$, where each element $\widetilde{A}_{mnij}\!\!=\!1$ if the $i$-th  value of the $m$-th entity in $G$ corresponds to the same attribute of the $j$-th value of the $n$-th entity in $G'$, and $\widetilde{A}_{mnij}\!\!=\!0$ otherwise. 
Then we calculate the element-wise product of $\widetilde{\mathbf{V}}$ and $\widetilde{\mathbf{A}}$, i.e., $\widetilde{\mathbf{V}} \odot \widetilde{\mathbf{A}}$, to get the masked value similarity matrix.
Finally, we summarize the similarities of all the $M^2$ attribute pairs for each entity pair to obtain an  entity similarity matrix: 
\beq{
	\label{eq:entity_similarity}
		\mathbf{S}^\mathcal{A}_e = \sum_{i}^{M} \sum_{j}^{M} \widetilde{V}_{\cdot, \cdot, i,j} \odot \widetilde{A}_{\cdot, \cdot, i,j},
\vspace{-0.1in}
}
\noindent 
$\mathbf{S}^\mathcal{A}_e \in \mathbb{R}^{N \times N'}$.
The superscript $^\mathcal{A}$ indicates the entity similarities are estimated by the attribute model.
The above matrix computation is quite efficient, as the most expense comes from the construction of the value embedding matrices, which only requires invoking the translation model $O(N \times M)$ times.

\subsection{Embedding-based Relationship Model}

Due to the success of the existing works on modeling the structures of the graph comprised by $\{( h, r, t )\}$~\cite{sun2018bootstrapping,zhang2019multi}, we adopt
TransE algorithm to maximize the energy (possibility) that  $h$ can be translated to $t$ in the KG, i.e., $E(h,r,t) = ||\mathbf{h} + \mathbf{r} - \mathbf{t}||$,  where $\mathbf{h}$, $\mathbf{r}$  and $\mathbf{t}$ represent the structure embeddings.

To preserve the cross-lingual relations of entities and relationships included in the existing alignments, we swap the entities or relationships in each alignment $(e\sim e')$ or $(r \sim r')$ to generate new relationship triplets~\cite{sun2018bootstrapping}. 
Then a margin-based loss function is optimized on the all relationship triplets to obtain entity embeddings $\mathbf{H}_e \in \mathbb{R}^{N \times D_e}$, $\mathbf{H}'_e \in \mathbb{R}^{N' \times D_e}$ and relationship embeddings $\mathbf{H}_r \in \mathbb{R}^{L \times D_r}$ and $\mathbf{H}'_r \in \mathbb{R}^{L' \times D_r}$ for both $G$ and $G'$, where $L$ and $L'$ are the number of relationships. and $D_e$ and $D_r$ are the embedding sizes with $D_e=D_r$.

\subsection{Jointly Modeling the Attribute and Relationship Model}
\label{sec:co-train}
Different from existing works that  learn combined embeddings by the attribute and the relationship model~\cite{sun2017crosslingual,Trsedya2019,zhang2019multi}, we propose a joint framework to firstly infer the confident alignments by the two models and then combine their inferences by three different merge strategies. Algorithm~\ref{algo:overview} illustrates the whole process. At each iteration, for modeling the attribute triplets, we first train the translation model based on the seed set of the aligned values (Line 3). Then we construct the value embedding matrices by the translation model (Line 4) and meanwhile construct the mask matrices by the existing aligned attributes (Line 5), based on which we perform an efficient matrix-based strategy to calculate the entity similarities (Line 6) and finally infer the new alignments of entities, attributes, and values based on the estimated similarities and existing alignments (Line 7). For modeling the relationship triplets, 
we train the entity and relationship embeddings based on the swapped relationship triplets between two graphs (Line 8), then we infer the new alignments of entities and relationships (Line 9). 
Finally, we merge the new aligned entity seeds from the attribute and the relationship model (Line 10) and augment the seed set by all the new alignments (Line 11 and 12). The framework bootstraps the two models iteratively by the extended alignments. Note we remove the new alignments from the candidate pairs at each iteration to avoid duplicate inference (Line 13). 

\begin{algorithm}[t]
	\caption{\model\label{algo:overview}}
	\scriptsize 
	\KwIn{$G$, $G'$ and the seed alignments $I = \{(e\sim e')\} \cup \{(r \sim r')\} \cup \{(a \sim a')\} \cup \{(v \sim v')\}$.}
	\KwOut{ The augmented alignments $I$.}
	$C$ = all the candidate pairs except the seed pairs in $I$;\\
	\Repeat{$ \triangle I = \emptyset$}{
		\tcc{Attribute model}
		Train a translation model  on $\{ (v \sim v')\}$;\\
		Construct the value embedding matrices $\textbf{V}$ and $\textbf{V}'$ by the translation model;\\
		Construct the mask matrices $\textbf{A}$ and $\textbf{A}'$ by $\{(a \sim a')\}$; \\
		Calculate entity similarities $\mathbf{S}^\mathcal{A}_e$ by Eq.~\eqref{eq:entity_similarity};\\
		Infer new alignments $I_e^\mathcal{A}$, $I_a$, $I_v$ from $C$ by Eq.\eqref{eq:infer_entity},\eqref{eq:infer_attribute},\eqref{eq:infer_value};\\
		
		\tcc{Relationship model}
		
		Train the relationship model on $\{(h,r,t)\}$, $\{(h', r',t')\}$, $\{(e \sim e')\}$ and $\{( r \sim r') \}$ to obtain $\mathbf{H}_e$, $\mathbf{H}'_e$, $\mathbf{H}_r$ and $\mathbf{H}'_r$;\\
		Infer new alignments $I_e^\mathcal{R}$ and $I_r$ from $C$ by Eq.~\eqref{eq:infer_relationship};\\
		
		\tcc{Merge new alignments}
		
		$I_{e} \leftarrow \text{merge} (I^\mathcal{A}_{e}, I^\mathcal{R}_{e})$; \\
		$\triangle I \leftarrow (I_e, I_r, I_a, I_v)$;\\
		$I = I + \triangle I$; 	\\
		$C = C - \triangle I$;\\
	}
	\normalsize
	
\end{algorithm}

\vpara{Infer Alignments by the Attribute Model.}
We select an entity pair $(e_m \!\sim\! e'_n)$ from  all the $N \times N'$ candidate entity pairs into the new aligned set of entities $I_e^\mathcal{A}$ if their similarity $S_e^\mathcal{A}[m,n]$ is larger than a threshold $\tau_e^\mathcal{A}$:
\beq{
	\label{eq:infer_entity}
	I_e^\mathcal{A} \leftarrow (e_m \sim e'_n),  \text{          if           } S_e^\mathcal{A}[m,n] > \tau_e^\mathcal{A}.
}
The candidates of the aligned attributes and values depend on the aligned entities.  Specifically, for each aligned entity pair $(e_m \!\sim \!e'_n)$, if the similarity of a value pair $\widetilde{V}_{m,n,i,j} $ is larger than a threshold $\tau_v$, we select their corresponding attribute pair $(a_i\! \sim \!a'_j)$ into the new aligned attribute set $I_a$: 
\beq{
	\label{eq:infer_attribute}
	I_a \leftarrow (a_i \sim a'_j), \forall (e_m \sim e'_n) \in I, \text{      if        } \widetilde{V}_{m,n,i,j} > \tau_v.
}
Then for each pair of attribute triplets  $(e_m,a_i,v_i) \in G$ and $(e'_n,a'_j,v'_j) \in G'$, if the entities and the attributes are both aligned, we select their corresponding  value pairs $(v_i \sim v'_j)$ into the new aligned value set $I_v$: 
\beqn{
	\label{eq:infer_value}
	I_v \!\leftarrow(v_i \sim v'_j), \!\!&\forall&\!\! (e_m,a_i,v_i) \in G \text{ \& }  (e'_n,a'_j,v'_j) \in G',\\\nonumber
	\!\!\!\!\!\!&\text{if}\!\!&  (e_m \sim e'_n) \in I  \text{   \&  } (a_i \sim a'_j) \in I. \nonumber
}

\vpara{Infer Alignments by the Relationship Model.}
We calculate the similarity matrix $\mathbf{S}_e^\mathcal{R}$ as the dot product of the entity embeddings where the superscript $^\mathcal{R}$ indicates the entity similarities are estimated by the relationship model.  Then we select an entity pair $(e_m \sim e'_n)$ into the new aligned entity set $I_e^\mathcal{R}$ if their similarity $S_e^\mathcal{R}[m,n]$ is larger than a threshold $\tau_e^\mathcal{R}$:
\beq{
	\label{eq:infer_relationship}
	I_e^\mathcal{R} \leftarrow (e_m \sim e'_n),  \text{          if           } S_e^\mathcal{R}[m,n] > \tau_e^\mathcal{R}.
}

The new aligned relationships are inferred in the same way but with a different threshold $\tau_r$.

\hide{
	\beq{
		I \leftarrow (r_m \sim r'_n),  \text{          if           } S_r[m,n] > \tau_r,
	}
	
	\noindent where $\tau_r$ is set heuristically, as the validate set about the aligned relationships are not available. One-one-mapping constraint is also applied.
	
}

\vpara{Merge Alignments of the Two models.}
We propose three strategies to merge $I_e^\mathcal{A}$ and $I_e^\mathcal{R}$ into $I_e$.

\ipara{Standard Multi-view Merge Strategy.}
Following the standard co-training algorithm, we firstly infer $I_e^\mathcal{A}$ from candidate entity pairs $C_e$ by Eq.~\eqref{eq:infer_entity}. 
Then we remove $I_e^\mathcal{A}$ from $C_e$, and then infer $I_e^\mathcal{R}$ from the remaining candidates $C_e-I_e^\mathcal{A}$. 

\ipara{Score-based Merge  Strategy.}
Due to the missing attributes and relationships, the labels inferred from the two views may have conflicts. 
For all the conflicting counterparts of an entity $e_m$, i.e., $\mathcal{C}_m = \{ e'_n | (e_m \sim e'_n) \in  I_e^\mathcal{A}\cup I_e^\mathcal{R}\}$, we select the counterpart with the maximal score $S_e[m,n] = S_e^\mathcal{A}[m,n] + S_e^\mathcal{R}[m,n]$ into the final new alignments. The strategy assumes that the alignments discovered by more views will be more confident:
\beq{ 
	I_e \leftarrow (e_m \sim e'_n), e'_n = \text{argmax}_{ e'_n \in \mathcal{C}_m} S_e[m,n].
 }

\ipara{Rank-based Merge  Strategy.}
Directly comparing the similarities estimated by the two models may suffer from the different scales of scores. Thus, we compare the normalize ranking indexes of the conflicting alignments. Specifically, for all the conflicting counterparts $\mathcal{C}_m$ of $e_m$, we select the counterpart with the minimal ranking ratio $R[m,n]$ into the final new alignments:
\beqn{
	\nonumber
	I_e &\leftarrow& (e_m \sim e'_n), e'_n = \text{argmin}_{ e'_n \in \mathcal{C}_m} R[m,n], \\ 
	\nonumber
	R[m,n]&=& \left
	\{\begin{array}{cl} 
		r^\mathcal{A}[m,n]/|I_e^\mathcal{A} |, & \text{if }  (e_m \sim e'_n) \in  I_e^\mathcal{A};   \\
		r^\mathcal{R}[m,n]/|I_e^\mathcal{R} |,  & \text{if } (e_m \sim e'_n) \in  I_e^\mathcal{R}.  
	\end{array}\right.  \nonumber
}
\noindent where $r^\mathcal{A}[m,n]$, $r^\mathcal{R}[m,n]$ denote the ranking index of the alignment $(e_m \!\sim \!e'_n)$ in $I_e^\mathcal{A}$ and $I_e^\mathcal{R}$ respectively, and $R[m,n]$ denotes the normalized ranking index. 

%% file: exp.tex
\begin{table}
	\newcolumntype{?}{!{\vrule width 1pt}}
	\newcolumntype{C}{>{\centering\arraybackslash}p{3.8em}}
	\newcolumntype{D}{>{\centering\arraybackslash}p{3.6em}}
	\caption{
		\label{tb:dataset} Data statistics. \small{Notation \#Rt denotes the number of the relationship triplets, \#Ar denotes the number of the attribute triplets.}}
	\centering \scriptsize
	\renewcommand\arraystretch{1.0}
	\begin{tabular}{c@{~}?c*{1}{CCCCC}@{~}?}
		\toprule
		Dataset & {\#Ent.} & {\#Rel.} & {\#Attr.} & {\#Rt} & {\#At}  \\
		\midrule
		ZH-EN  &   164,594 &	5,147&	15,286	& 391,603 &	947,439  \\
		 JA-EN  & 161,424	& 4,139 &	11,948	& 397,692 &	851,849\\ 
		FR-EN  & 172,747 &	3,588 &	10,969	& 470,781 &	1,105,208 \\

		\bottomrule
	\end{tabular}
	\normalsize
\end{table}

\section{Experiments}
\label{sec:exp}

\subsection{Experimental Settings}

\vpara{Dataset.}
We evaluate the proposed model on DBP15K\footnote{https://github.com/nju-websoft/JAPE}, a well-known public dataset for KG alignment. DBP15K contains 3 pairs of cross-lingual KGs, each of which contains 15,000 inter-lingual links (ILLs). The proportion of the ILLs for training, validating and testing is 4:1:10.  Table~\ref{tb:dataset} shows the data statistics.

\vpara{Baseline Methods.}  We compare several existing methods:

\textit{MuGNN}~\cite{cao2019multi}: Learns the structure embeddings by a multi-channel GNNs.

\textit{BootEA}~\cite{sun2018bootstrapping}: Is  a bootstrap method that finds new alignments by performing a maximal matching between the structure embeddings of the entities.

\textit{JAPE}~\cite{sun2017crosslingual}: Leverages the attributes and the type of values  to refine the structure embeddings.

\textit{GCNs}~\cite{wang2018cross}: Learns the structure embeddings by GCNs and use the one-hot representation of all the attributes as the initial input of an entity.


\textit{MultiKE}~\cite{zhang2019multi}: Learns a global attribute embedding for each entity and combines it with the structure embedding. Since it is to solve monolingual entity alignment, for a fair comparison, we translate all the words into English by Google's translator and then applies MultiKE.


\textit{\model:} Is our model. The variant \samodel removes the relationship model and \srmodel removes the attribute model, the bootstrap strategy is still adopted. 


As KDCoE~\cite{chen2018co-training} and Yang et al.~\cite{yang2019aligning} leverage the descriptions of entities, and Xu et al.~\cite{xu2019cross} adopts external cross-lingual corpus to train embeddings, we do not compare with them and leave the studies with these resources in the future.


\vpara{Evaluation Metrics.}
In the test set, for each entity in $G$, we rank all the entities in $G'$ by either $\mathbf{S}_e^\mathcal{A}$ or $\mathbf{S}_e^\mathcal{R}$, and evaluate the ranking results by HitRatio@K (HRK), i.e., the percentage of entities with the rightly aligned entities ranked before top K, and Mean Reciprocal Rank (MRR), i.e., the average of the reciprocal ranks of the rightly aligned entities. 

\vpara{Implementation Details.}
In the attribute model, the  value embedding size $D_v$ is 100, the maximal number of attributes $M$ is 20, and the frequent attributes are those occurred more than 50 times in $\{(h,a,v)\}$. In the relationship model, the entity/relationship embedding size $D_e$ or $D_r$ is 75, 
and $\gamma = 1.0$.
The thresholds $\tau^A_e$ and $\tau^R_e$ for selecting the aligned entities are set as the values when the best HR1 is obtained on the validation set. $\tau_v$ for selecting the aligned attributes is 0.8, and $\tau_r$ for selecting the aligned relationships is 0.9. 

\vpara{Initial Seeds Construction.}
The existing ILLs can be viewed as the entity seed alignments.  
Some relationships or attributes in cross-lingual knowledge graphs are both represented in English. So we can easily treat a pair of relationships or attributes with the same name\footnote{Attributes and relationships are less ambiguous than entities. The sampled 500 pairs of attributes and relationships present that about 95\% of them can be safely aligned only based on the same names.} as a relationship or attribute seed alignment.
Finally, the corresponding values of the aligned attributes for any aligned entity pairs are added into the seed set of the aligned values.

\subsection{Experimental Results}

\begin{table*}
	\newcolumntype{?}{!{\vrule width 1pt}}
	\newcolumntype{C}{>{\centering\arraybackslash}p{3.0em}}
	\caption{
		\label{tb:performance} Overall performance of entity alignment (\%).
		\normalsize
	}
	\centering \scriptsize
	\renewcommand\arraystretch{1}
	\setlength{\tabcolsep}{0.1mm}{ 
	\begin{tabular}{@{~}l@{~}?*{1}{CCC?}*{1}{CCC?}*{1}{CCC}@{~}}
		\toprule
		
		\multirow{2}{*}{\vspace{-0.3cm} Model}
		&\multicolumn{3}{c?}{DBP15K$_{\text{ZH-EN}}$}
		&\multicolumn{3}{c?}{DBP15K$_{\text{JA-EN}}$} 
		&\multicolumn{3}{c}{DBP15K$_{\text{FR-EN}}$}
		\\
		\cmidrule{2-4} \cmidrule{5-7} \cmidrule{8-10}
		& {HR1} & {HR10} & {MRR} & {HR1} & {HR10} & {MRR} & {HR1} & {HR10} & {MRR} \\
		\midrule
		MuGNN 
		& 49.40	& 84.40	& 61.10
		& 50.10	& 85.70	& 62.10
		& 49.50	& 87.00	& 62.10
		\\
		BootEA
		& 62.94 & 84.75	& 70.30
		& 62.23	& 85.39	& 70.10
		& 65.30 & 87.44	& 73.10
		\\ 
		JAPE 
		& 41.18	& 74.46	& 49.00
		& 36.25	& 68.50	& 47.60
		& 32.39	& 66.68	& 43.00
		\\
		GCNs 
		& 41.25 & 74.38 & 55.80
		& 39.91 & 74.46	& 55.20
		& 37.29 & 74.49	& 53.40
		\\
		MultiKE
		&50.87 &57.61 &53.20	
		&39.30 &48.85	&42.60
		&63.94 &71.19	&66.50
		
		\\\midrule
		\samodel
		& 57.18 & 70.44 & 61.80
		& 50.63	& 60.36	& 54.30
		& 53.92	& 60.40	& 56.30
		\\		
		
		\srmodel
		& 58.64	& 83.89	& 67.10
		& 55.74	& 83.23	& 65.10
		& 59.25	& 85.74	& 68.60
		\\\midrule		
		
		\model (M1)
		&68.59 &86.56	&74.90
		&62.65 &82.79	&69.70
		&68.43 &87.86	&75.10
		\\
		\model (M2)
		&69.32 &87.37	&75.50
		&63.01 &83.37	&70.00
		&\textbf{70.87} &87.05	&76.50
		\\
		\model (M3)
		& \textbf{70.58}	& \textbf{87.81}	& \textbf{76.60}
		& \textbf{64.58}	& \textbf{85.50}	& \textbf{70.80}
		& 70.41	& \textbf{88.81}	& \textbf{76.80}
		\\\midrule
		\model-IT
		&66.39 &87.29	&73.40
		&60.08 &84.45	&68.20
		&68.31 &88.33	&75.40
		\\\bottomrule
	\end{tabular}
}
\end{table*}

\vpara{Overall Alignment Performance.}
Table \ref{tb:performance} shows the overall performance of entity alignment. 
MuGNN and BootEA only leverage relationship triplets. BootEA bootstraps the alignments iteratively and performs better than MuGNN. 
Although JAPE and GCN additionally consider the attribute triplets, they perform much worse than BootEA, as they only leverage the attributes but ignore their corresponding values.  
MultiKE utilizes the values and performs better than JAPE and GCN. 
However, it learns and compares the global embeddings of entities, which may bring in additional noises by the irrelevant attribute triplets. 

\begin{figure}
	\centering
	
	\subfigure[Effect of $\tau_v$]{\label{subfig:atm}
		\includegraphics[width=0.4\textwidth]{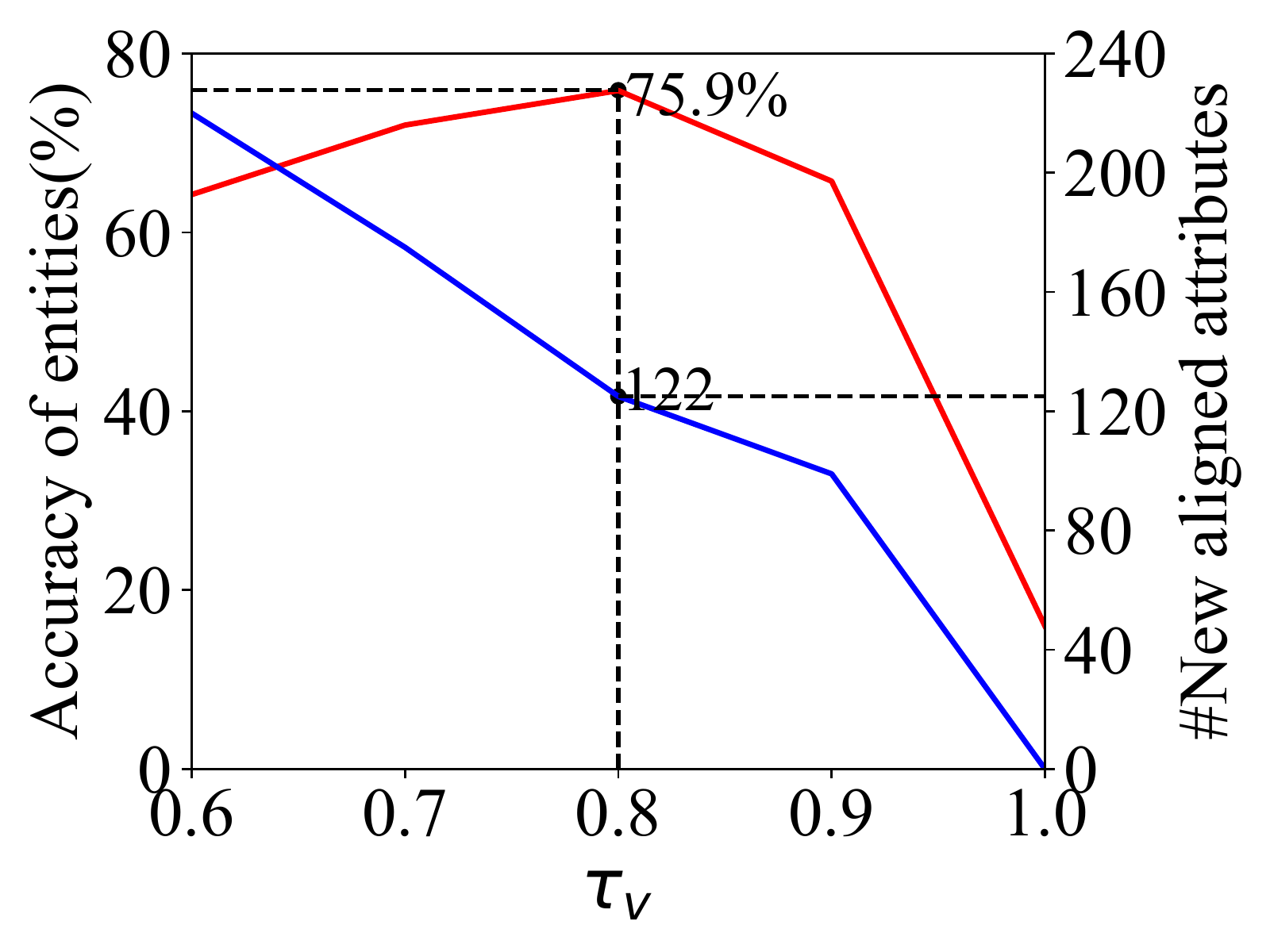}
	}
	\centering
	\subfigure[Case study]{\label{subfig:casestdy}
		\includegraphics[width=0.445\textwidth]{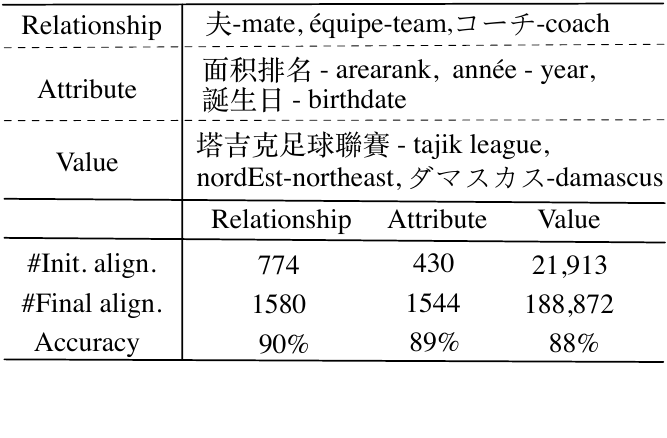}
	}
	
	\caption{\label{fig:analysis} Parameter Analysis and Case Study.}
\end{figure}

\smodel proposes an interaction-based attribute model which directly compares the values of the aligned attributes, thus it clearly  performs better than others (+2.35-38.48\% in HR1).  
\smodel also outperforms the variant \samodel and \srmodel. Specifically, \samodel is comparable to \srmodel in HR1 but underperforms in HR10 and MRR. Because in \samodel, we set a strict threshold (Cf. Fig.~\ref{subfig:atm}) to obtain high-qualified alignments, which makes the translation model not easy to include the difficult alignments into the training data, i.e.,  the seemingly irrelevant value pairs which in fact indicate the same things. 


\vpara{The Effect of Different Merge Strategies.}
We compare the effects of the proposed three merge strategies and show the results of \model(M1), (M2) and (M3) in Table~\ref{tb:performance}. We can see that the standard multi-view merge strategy(M1) performs worst, as it does not solve the conflicts from the two views. The score-based merge strategy (M2) and the rank-based merge strategy (M3) solve the conflicts, thus perform better than M1 (+1.26-2.43\% in HR1). M3 avoids comparing the scores of different scales, thus performs better than M2 in most of the metrics. Later, \smodel indicates the proposed model with M3.

\vpara{The Effect of Iteratively Update the Translation Model.}
We validate the effect of \underline{i}teratively updating the \underline{t}ranslation model (IT) during the joint modeling process. Specifically,  we compare \smodel with the translation model being trained only once at the beginning, which is denoted as \model-IT. From Table~\ref{tb:performance}, we can see that \model-IT performs worse \smodel by 2.56-4.50\% in HR1, which indicates that the newly discovered value alignments by our model can boost the performance of  the translation model.


\vpara{The Effect of $\tau_v$ and $\tau_r$.} 
We verify how the new aligned attributes can benefit the entity alignment. Specifically, we vary the threshold $\tau_v$ from 0.6 to 1.0 with interval 0.1 and show the results of \samodel on DBP15K-1 FR-EN in Fig.~\ref{subfig:atm}. It is shown that when $\tau_v\!=\!1.0$, i.e., \#new aligned attributes is 0, the accuracy of entity alignment is significantly hurt. 
When $\tau_v \!<\! 1.0$, with the increase of \#new aligned attributes, the accuracy improves and approaches the best when $\tau_v\!=\!0.8$, as the quantity and the quality of the new aligned attributes are well balanced. The threshold $\tau_r$ for finding the new aligned relationships is set in the same way.

\vpara{Case Study.}
We present several cases of the new aligned relationships, attributes and values in different languages by \smodel on DBP15K in Fig.~\ref{subfig:casestdy}. We also show the number of initial and the finally discovered alignments on DBP15K ZH-EN. Most of the newly discovered alignments are high-frequent attributes or relationships. The low-frequent attributes or relationships are difficult to be aligned by the current method and will be studied in the future.
We randomly sample 100 final alignments and manually evaluate the accuracy, as their ground truth is not available. The results demonstrate the effectiveness of our model. The whole alignments together with the codes are available online\footnote{https://github.com/BoChen-Daniel/PAKDD-20-JarKA}.

%% file: conclusion.tex
\section{Conclusions and Future Work}
\label{sec:con} 

We present the first attempt to formalize the problem of cross-lingual knowledge alignment  as comprehensively linking entities, relationships, attributes and values. We propose an interaction-based attribute model to compare the aligned attributes of entities instead of globally embedding entities. A matrix-based strategy is adopted to accelerate the comparing process.  Then we propose a joint framework together with three merge strategies to solve the conflicts of the alignments inferred from the attribute model and the relationship model.
The experimental results demonstrate the effectiveness of the proposed model. In the future, we plan to  incorporate the descriptions of entities and the pre-trained cross-lingual language model to enhance our model's performance.

\small
\vpara{Acknowledgments.} This work is supported by National Key R\&D Program of China (No.2018YFB1004401) and NSFC under the grant No. 61532021, 61772537, 61772536, 61702522.
*Jing Zhang is the corresponding author.
\normalsize